\documentclass[letterpaper]{article} 
\usepackage{aaai2026}  
\usepackage{times}  
\usepackage{helvet}  
\usepackage{courier}  
\usepackage[hyphens]{url}  
\usepackage{graphicx} 
\usepackage{natbib}  
\usepackage{caption} 
\usepackage{makecell}
\usepackage{algorithm}
\usepackage{algorithmic}
\usepackage{amsmath}
\usepackage{xcolor}
\usepackage{graphicx}
\usepackage{multirow}
\usepackage{booktabs}
\usepackage{adjustbox}
\usepackage{colortbl}
\usepackage{xcolor}
\usepackage{amssymb}
\usepackage{enumitem}
\usepackage{newfloat}
\usepackage{listings}
\DeclareCaptionStyle{ruled}{labelfont=normalfont,labelsep=colon,strut=off} 
\floatstyle{ruled}
\newfloat{listing}{tb}{lst}{}
\floatname{listing}{Listing}
\newcommand{\hide}[1]{}
\newtheorem{theorem}{Theorem}
\newtheorem{definition}{Definition}

\title{MoKA: Mixture of Kronecker Adapters}
\author{
    Mohammadreza Sadeghi\textsuperscript{\rm 1}, Mahsa Ghazvini Nejad\textsuperscript{\rm 1}, MirHamed Jafarzadeh Asl\textsuperscript{\rm 1}, Yu Gu\textsuperscript{\rm 1,2}, Yuanhao Yu\textsuperscript{\rm 1}, Masoud Asgharian\textsuperscript{\rm 2}, Vahid Partovi Nia\textsuperscript{\rm 1}
}
\affiliations{
    \textsuperscript{\rm 1}Huawei Noah's Ark Lab\\
    \textsuperscript{\rm 2}Department of Mathematics and Statistics, McGill University\\

}

\usepackage{bibentry}

\begin{document}

\maketitle

\begin{abstract}
\hyphenpenalty=100000
\exhyphenpenalty=100000
Parameter-efficient fine-tuning (PEFT) is essential for reducing the computational overhead of large language models (LLMs). Low-rank family adapters are commonly used to control the parameter size efficiently while maintaining the generative power of LLMs. However, their limited expressiveness due to the rank constraint often restricts their performance on complex tasks.
We propose Mixture of Kronecker Adapters~(MoKA), a new generation of Kronecker adapters that addresses this limitation by modeling weight updates as a mixture of Kronecker products.
Our proposed adapter leverages a gating mechanism that measures the importance of each Kronecker factor, enabling more expressive adaptation. Moreover, MoKA enables a rank flexibility that provides a better trade-off between parameter efficiency and accuracy. 
To ensure hardware efficiency, we reformulate Kronecker computations using standard matrix operations, allowing seamless deployment on GPU-optimized hardware. We conduct extensive experiments on instruction-tuning and commonsense reasoning tasks using low-bit quantized versions of LLaMA2-7B and LLaMA3-8B models. MoKA not only outperforms PEFT baselines, but also reduces the number of trainable parameters up to $\approx 27\times$, achieving state-of-the-art trade-offs between performance and parameter efficiency.
\end{abstract}

\section{Introduction}
\label{sec:introduction}
Parameter-efficient fine-tuning (PEFT) offers a practical way to adapt large language models (LLMs) without the heavy computational cost of full-model updates. By updating only a small subset of parameters, often introduced as lightweight modules such as low-rank adapters~\cite{lora, qlora, dora}, prefix-tuning vectors~\cite{li2021prefix}, or other structural augmentations, PEFT~\cite{shi2024reference} significantly reduces memory footprint and training time while retaining competitive performance. As a result, PEFT is especially valuable in resource-constrained settings or when adapting LLMs to a variety of downstream tasks. 

Among PEFT methods, low-rank adapters, such as LoRA~\cite{lora}, DoRA~\cite{dora}, and QLoRA~\cite{qlora} are widely adopted and studied due to their simplicity and effectiveness in reducing parameter count. These approaches typically approximate weight updates using low-rank matrix decompositions, which significantly limits the number of trainable parameters. However, since the low-rank constraint inherently restricts the expressiveness of the parameter space, it is challenging for the model to capture complex task-specific patterns. As a result, while low-rank adapters perform well on many tasks, they often struggle to adapt to difficult or more nuanced tasks where richer parameterization is required for effective fine-tuning.

\begin{figure}[t]
    \centering
    \includegraphics[width=0.95\columnwidth]{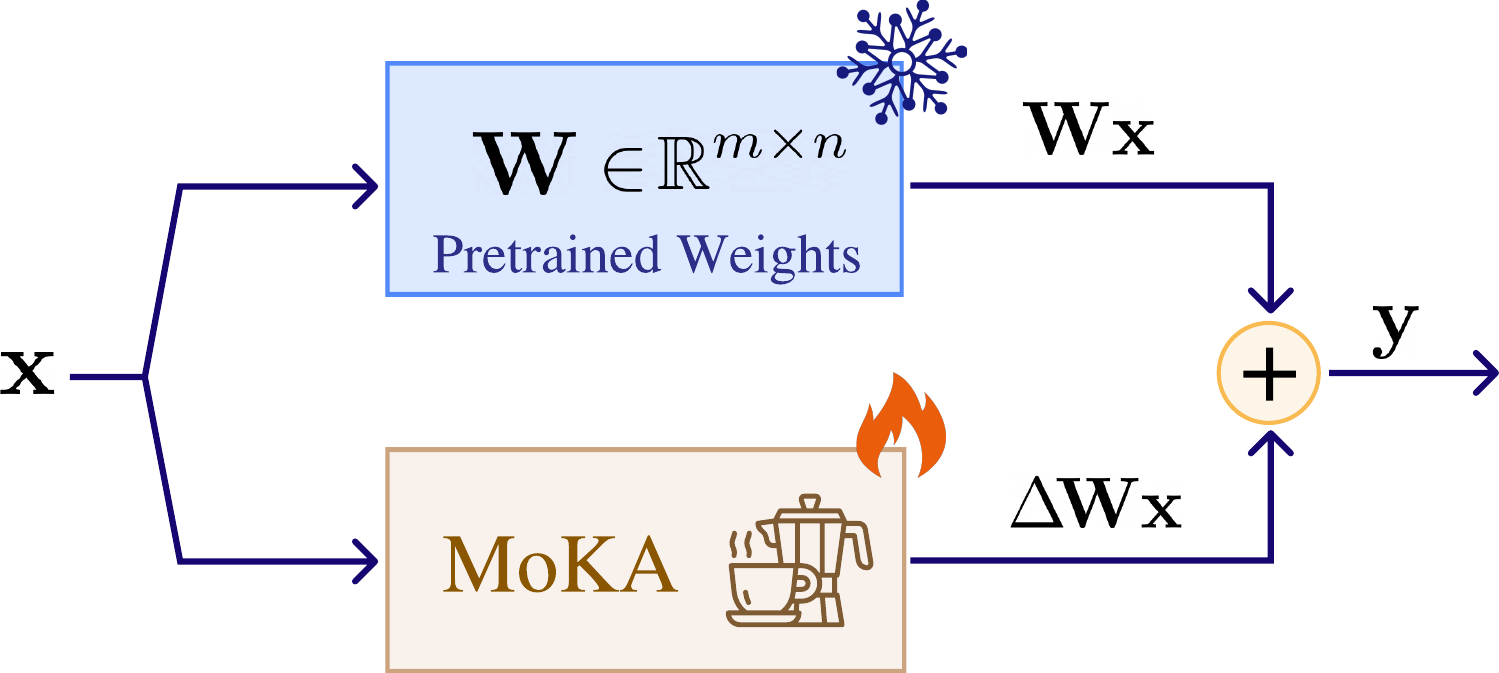}
    \caption{An overview of the MoKA pipeline. The layer weights of the pretrained model are reparameterized using the MoKA module. Here, $\mathbf{x}$ denotes the layer input and $\mathbf{y}$ represents the linear output before the nonlinear activation. The pretrained weights remain frozen, while only the MoKA parameters are trained.}
    \label{fig:pipeline}
\end{figure}

To address the limitations of low-rank adaptation, Kronecker-product-based adapters \cite{Krona, braga2024adakron} have been proposed as a more expressive alternative. By modeling weight updates using Kronecker factorization, these adapters offer a higher-capacity parameterization without significantly increasing the number of trainable parameters. However, they have not seen widespread adoption in practice. One key limitation is that Kronecker decomposition imposes a structural assumption on the adapted weights, which may not align with the optimal update patterns required for specific tasks. Additionally, modern hardware and deep learning frameworks are primarily optimized for dense matrix operations, not for Kronecker products. This lack of hardware support makes the computation of Kronecker-based updates more complex and potentially inefficient, hindering their practical applicability despite their theoretical advantages.

\begin{figure*}[t]
    \centering
    \includegraphics[width=0.85\textwidth]{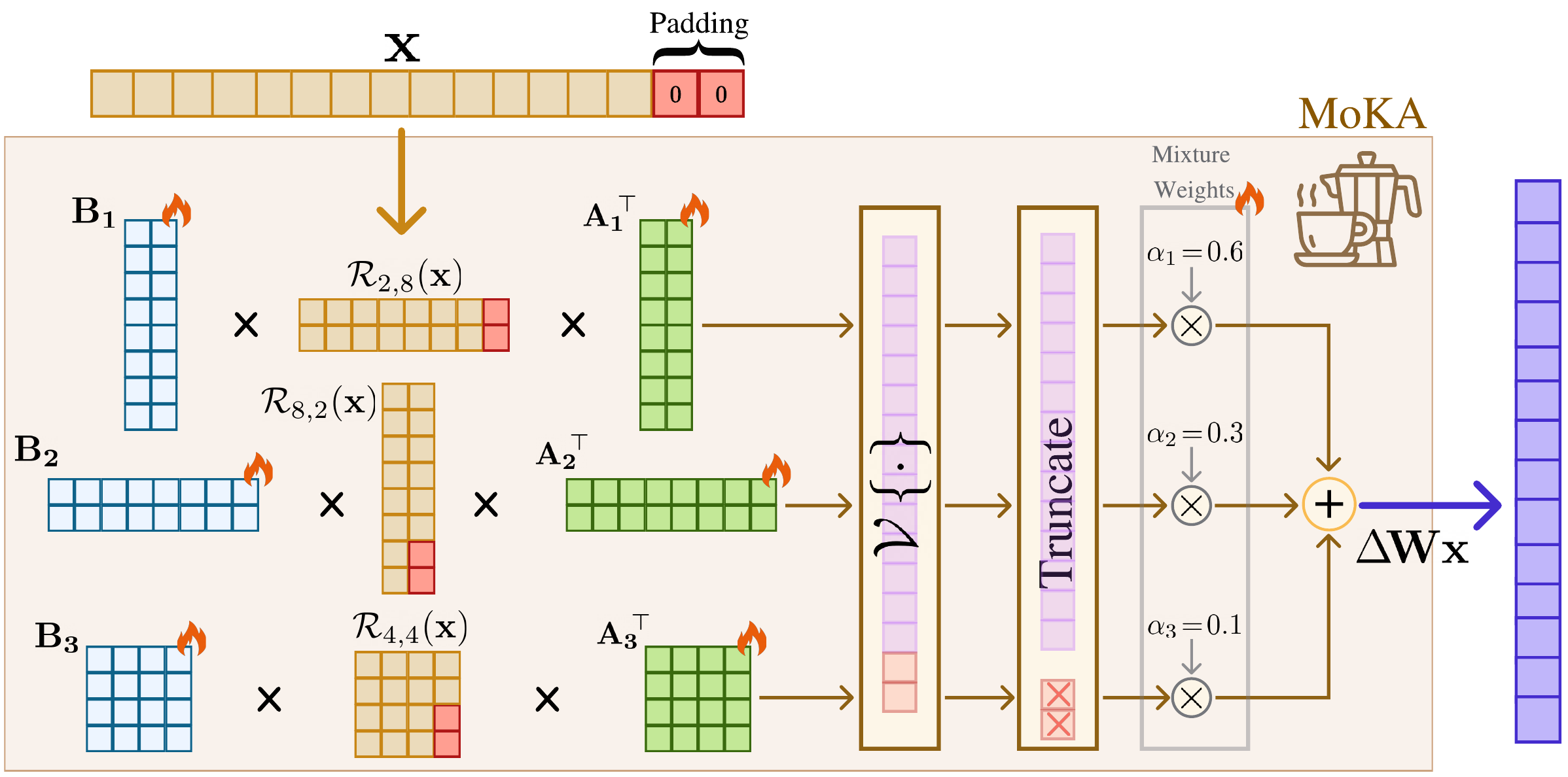}
    \caption{An overview of the MoKA architecture. Multiple sets of Kronecker filters are used to effectively adapt the model to downstream tasks. The input $\mathbf{x}$ is padded and reshaped to match the required dimensions for matrix multiplication. The learnable parameters of MoKA include the Kronecker filters ($\mathbf A_i$ and $\mathbf B_i$) and the mixture weights ($\alpha_i$).}
    \label{fig:moka}
\end{figure*}

In this paper, we propose Mixture of Kronecker Adapters (MoKA), a new generation of Kronecker-based PEFT that addresses both the structural and computational limitations of prior Kronecker adapters. Instead of relying on a single Kronecker factorization, MoKA combines multiple Kronecker adapters with diverse filter shapes through a learnable gating mechanism that dynamically weighs the contribution of each adapter based on input context. This mixture formulation significantly expands the expressiveness of the parameter space, enabling the model to capture a wider range of structural patterns without being confined to a fixed low-rank or rigid block structure. To make MoKA practical on modern hardware, we further introduce an efficient implementation trick that avoids explicit Kronecker multiplication, instead reducing all computations to standard matrix multiplications. This design allows MoKA to fully leverage highly optimized GPU kernels, making it both accurate and hardware-efficient for fine-tuning LLMs.
Our main contributions are as follows: 
\begin{itemize}
    \item We propose MoKA with diverse filter shapes to enhance representational capacity beyond traditional low-rank constraints.
    \item We introduce a learnable gating mechanism that adaptively selects and weighs Kronecker factors, allowing the model to flexibly capture varying structural patterns across tasks.
    \item We develop an efficient implementation that avoids explicit Kronecker multiplications by reformulating computations in terms of standard matrix operations, enabling compatibility with GPU-optimized kernels.
    \item We evaluate MoKA on instruction-tuning and commonsense reasoning tasks using 4-bit quantized LLaMA2-7B and LLaMA3-8B models, showing that it outperforms existing PEFT methods while cutting trainable parameters by up to 27$\times$, achieving a better trade-off between performance and efficiency.
    \item Our theoretical convergence study shows that Kronecker structure follows convergence properties similar to unconstrained fine-tuning.
\end{itemize}



\section{Related Work}
\label{sec:related_work}
PEFT methods have evolved through time to adapt LLMs efficiently. {Adapter Tuning}~\cite{houlsby2019parameter} is one of the first PEFT methods, which introduces task-specific feed-forward modules with a bottleneck structure into transformer layers. This method achieves near full fine-tuning performance with fewer parameters. However, its fixed architecture limits flexibility across diverse tasks. Subsequently, {Prefix Tuning}~\cite{lester-etal-2021-power} optimizes continuous task-specific tokens prepended to inputs, leveraging existing weights. Nevertheless, its reliance on non-semantic tokens demands extensive training for optimal results. {BitFit}~\cite{ben-zaken-etal-2022-bitfit} simplifies PEFT by fine-tuning only bias terms. Although this approach reduces computational costs, it lacks the capacity for complex adaptations due to limited parameter scope.

A major advancement in PEFT appeared with {LoRA}~\cite{lora}, which decomposes weight updates into low-rank matrices, minimizing the number of trainable parameters while preserving performance. A key practical limitation is the need for manual rank selection, which varies by task and requires tuning~\cite{lora}.
Building on this, {DyLoRA}~\cite{valipour2022dylora} dynamically selects ranks during training, enhancing adaptability through a nested dropout-inspired approach. However, its iterative rank sampling increases computational costs compared to LoRA~\cite{Wang2025-PEFTsurvey}.
{DoRA}~\cite{dora} decomposes pretrained weights into magnitude and direction, applying {LoRA} solely to directional updates. Without extra inference overhead, {DoRA} improves the learning capacity and training stability of {LoRA}. However, this technique struggles with tasks that require significant weight restructuring due to its constrained adaptation scope.
Furthermore, {AdaLoRA}~\cite{adalora} adaptively allocates parameters using Singular Value Decomposition (SVD) to prune less critical components, which enhances efficiency at the cost of added training complexity.

QLoRA~\cite{qlora} is a quantization-aware fine-tuning method that combines 4-bit quantization with {LoRA}, achieving near full-precision performance while reducing memory usage. Building on this, QDyLoRA~\cite{rajabzadeh-etal-2024-qdylora} introduces dynamic rank allocation within quantized models. Unlike {QLoRA}’s fixed-rank approach, {QDyLoRA} adapts rank per layer and task, allowing a single fine-tuned model to support multiple {LoRA} ranks and providing greater flexibility. Although both techniques increase the efficiency of {LoRA}, {QDyLoRA} prioritizes dynamic adaptability whereas {QLoRA} concentrates on memory optimization.

Among other methods, {UniPELT}~\cite{mao-etal-2022-unipelt}, a hybrid technique that integrates the advantages of multiple strategies, specifically adapters, prefix tuning, and LoRA. However, this makes designing and hyperparameter tuning much more difficult. {AUTOPEFT}~\cite{zhou-etal-2024-autopeft}, an automated approach, uses multi-objective Bayesian optimization to streamline PEFT configuration selection, reducing manual effort but requiring significant computational resources.

Kronecker-based techniques, such as {KronA}~\cite{Krona} and {AdaKron}~\cite{braga2024adakron}, extend the concept of low-rank adaptation by incorporating Kronecker product-based decomposition.
In contrast to {LoRA}, which relies on simple low-rank matrices, {KronA} introduced Kronecker product-based adapters to parameterize weight updates, allowing for more expressive adaptations than {LoRA}, with a smaller parameter footprint.
Building on this work, {AdaKron} combines adaptive rank selection with structured decomposition. This technique achieves state-of-the-art performance in tasks that require high adaptability, such as text generation and multi-modal applications. However, its computational complexity during training remains a challenge for resource-constrained environments.


\section{Methodology}

\subsection{Background}
Numerical computations are  performed using Central Processing Units (CPUs) combined with additional accelerators such as Graphical Processing Units (GPUs), that are highly optimized to perform certain  operations such as sorting, searching, and matrix multiplication. A matrix multiplication is composed of several inner products, and an accelerator parallelizes this operation by assigning each inner product to a single or multiple computing cores. Deep learning models are composed of extensive matrix multiplication operations; therefore, using accelerators in AI deployment is inevitable.  Adapters are new tools to fine-tune an AI model without losing its generative capability. A commonly used method is to add trainable low-rank matrices to control the size of the learnable parameters. The low-rank structure is commonly realized through i) matrix inner product $\mathbf W_{m\times n}  = \mathbf A_{n\times r} \mathbf B _{r\times n}$ or ii) outer product.

The outer product reformulation of a low-rank structure is preferable in the context of parameter-efficient training. It allows us to interpret the rank $r$ as the flexibility parameter to approximate $\mathbf W $, \hide{and interpret the outer product  as the expansive operator to express $\mathbf W$,  }
$$\mathbf W_{m\times n} = \sum_{i=1}^r \mathbf a_i \mathbf b_i^\top,$$
where the vector $\mathbf a_i$ of length $m$ is the $i$th row of $\mathbf A,$ and the vector $\mathbf b_i^\top$ of length $n$ is the $i$th column of $\mathbf B$. The Kronecker approximation generalizes the outer product by exchanging the vector outer product $\mathbf a_i \mathbf b_i^\top$ with the matrix Kronecker product $\mathbf A_i \otimes \mathbf B_i$ where $\mathbf A_i$ and $\mathbf B_i$ are of size $m_a\times n_a$, and $m_b \times n_b$, respectively, $m= m_a m_b $ and $n=n_a n_b$, 
$$\mathbf W = \sum_{i=1}^r \mathbf A_i \otimes \mathbf B_i,$$ 
and 
\hide{It is easier to interpret Kronecker outer product generalization by writing the two operation explicitly as  
\begin{align}
\mathbf a  \mathbf b^\top = 
\begin{pmatrix}
a_{1}b_{1} & \cdots & a_{1}b_{n} \\
\vdots & \ddots & \vdots \\
a_{m}b_{1} & \cdots & a_{m}b_{n}
\end{pmatrix},
\label{eq:outer}
\end{align}
is the outer product} 
\begin{align}
\mathbf A \otimes \mathbf B = 
\begin{pmatrix}
a_{11}\mathbf B & \cdots & a_{1n_a}\mathbf B \\
\vdots & \ddots & \vdots \\
a_{m_a1}\mathbf B & \cdots & a_{m_an_a}\mathbf B
\end{pmatrix}.
\label{eq:kronecker}
\end{align}
\hide{is its Kronecker product generalization. }

We can rewrite Kronecker product projection of vector $\mathbf x$ in the form of matrix multiplication
\begin{align} 
(\mathbf A \otimes \mathbf B)\mathbf x = \mathcal{V}\left\{\mathbf B\, \mathcal{R}_{n_b,n_a}(\mathbf x)\, \mathbf A^\top\right\},
\label{eq:matirx_identity}
\end{align}
where $\mathcal{R}_{n_b,n_a}(\mathbf)$ reshapes the vector $\mathbf x$ into an ${n_b \times n_a}$ matrix, and $\mathcal {V}(\cdot)$ reshapes a matrix to a column vector, see~\cite{VanLoanPitsianis_KronApprox_1993}. The identity \eqref{eq:matirx_identity} allows to implement the Kronecker product through the matrix multiplication operation in which CPUs and GPUs are optimized for. This is a considerable advantage, because there is  no dedicated hardware to support the Kronecker product computation.

Several properties make the Kronecker product attractive for modeling weight matrices. First, dimensionality: if $\mathbf A$ and $\mathbf B$ are much smaller than the target matrix, $\mathbf A\otimes \mathbf B$ provides a compressed representation. Second, and importantly, a Kronecker factorization is not inherently rank-deficient. In fact, $\mathrm{rank}(\mathbf A \otimes \mathbf B) = \mathrm{rank}(\mathbf A) \times \mathrm{rank}(\mathbf B)$. Unlike a standard low-rank factorization which imposes a strict rank-$r$ constraint, a Kronecker decomposition can maintain high rank (even full rank) if the factors are appropriately chosen~\cite{Krona}. This means a Kronecker-modeled weight matrix can potentially express a richer family of transformations than a comparable low-rank model of similar size.

\subsection{Proposed Kronecker-Product Adapter Method} \label{sec:moka}

Let \( \mathbf{W} \in \mathbb{R}^{m \times n} \) denote a weight matrix from the pretrained model (e.g., a transformer's feed-forward or attention projection layer) that we aim to adapt for a downstream task. During fine-tuning, we modify this layer by introducing a task-specific residual component \( \Delta \! \mathbf{W} \), yielding the updated weight matrix:
\begin{align}
\mathbf{W}' = \mathbf{W} + \Delta \! \mathbf{W},
\end{align}
where \( \Delta \! \mathbf{W} \) is learned while keeping \( \mathbf{W} \) frozen. For an input vector \( \mathbf{x} \in \mathbb{R}^n \), the output of the adapted layer becomes:
\begin{align}
\mathbf{y} = \mathbf{W}' \mathbf{x} = (\mathbf{W} + \Delta \! \mathbf{W}) \mathbf{x} = \mathbf{W} \mathbf{x} + \Delta \! \mathbf{W} \mathbf{x}.
\end{align}

Here, \( \mathbf{W} \mathbf{x} \) retains the original knowledge from the pretrained model, while \( \Delta \! \mathbf{W} \mathbf{x} \) incorporates the task-specific adaptation. Fig.~\ref{fig:pipeline} shows an overview of the MoKA pipeline.

In \textbf{MoKA}, we model \( \Delta \! \mathbf{W} \) as a gated mixture of \( r \) Kronecker adapters as follow:
\begin{align}
\Delta \! \mathbf{W} = \sum_{i=1}^{r} \alpha_i \left( \mathbf{A}_i \otimes \mathbf{B}_i \right),
\end{align}
where \( \mathbf{A}_i \in \mathbb{R}^{m_{a_i} \times n_{a_i}} \) and \( \mathbf{B}_i \in \mathbb{R}^{m_{b_i} \times n_{b_i}} \) are learnable matrices. Each pair \( (\mathbf{A}_i, \mathbf{B}_i) \) represents a Kronecker adapter with a unique filter shape, allowing MoKA to explore a diverse set of matrix structures without being constrained to a fixed rank or pattern. The mixture weights \( \alpha_i \) are obtained via a softmax over a set of learnable gating parameters \( g_i \):
\begin{align}
\alpha_i = \operatorname{softmax}(g)_i = \frac{\exp(g_i)}{\sum_{j=1}^{r} \exp(g_j)}
\end{align}
This mixture design improves both flexibility and expressiveness while maintaining parameter efficiency.

Despite the advantages of MoKA, such as avoiding structural and low rank constraints, one notable drawback is the lack of hardware acceleration for Kronecker product operations. Specifically, no widely available hardware (e.g., GPUs or CPUs) offers support for efficient Kronecker computations, making direct evaluation of Kronecker products both memory-inefficient and computationally expensive. To this end, we leverage the identity in \eqref{eq:matirx_identity} to reformulate the Kronecker operation in terms of standard matrix multiplications and reshaping operations, which are highly optimized on modern hardware. Accordingly, we rewrite  \( \Delta \! \mathbf{W} \mathbf{x}  \) as follow:
\begin{align}\label{eq10}
    \Delta \! \mathbf{W} \mathbf{x} &= \sum_{i=1}^{r}\alpha_i(\mathbf A_i \otimes \mathbf B_i)\mathbf x \nonumber\\ & = \sum_{i=1}^{r}\alpha_i\mathcal{V}\left\{\mathbf B_i\, \mathcal{R}_{n_{b_i},n_{a_i}}(\mathbf x)\, \mathbf A_i^\top\right\},
\end{align}

As shown in \eqref{eq10}, the computation of each Kronecker adapter requires only reshaping and matrix multiplications, eliminating the need to explicitly form the Kronecker product. This formulation enables MoKA to remain compatible with matrix-multiplication-optimized hardware while retaining its expressive capacity. In our formulation, we do not impose specific constraints on the dimensions of the Kronecker factors \( \mathbf{A}_i \in \mathbb{R}^{m_{a_i} \times n_{a_i}} \) and \( \mathbf{B}_i \in \mathbb{R}^{m_{b_i} \times n_{b_i}} \).

In order to apply the reshape operation \( \mathcal{R}_{n_{b_i}, n_{a_i}}(\mathbf{x}) \), however, the input vector \( \mathbf{x} \in \mathbb{R}^{n} \) must satisfy \( n = n_{a_i} n_{b_i} \). To ensure this condition holds, we select Kronecker filter dimensions such that \( n_{a_i} n_{b_i} \leq n \). When this inequality is strict, that is \( n_{a_i} n_{b_i} < n \), we apply {\it zero padding} to the end of \( \mathbf{x} \), extending its length to exactly \( n_{a_i} n_{b_i} \). This padding operation ensures that the reshape is valid while preserving the original input content. After \eqref{eq10}, the resulting vector may exceed the desired output dimensionality. To resolve this, we truncate the excess entries from the end of the output, ensuring that the final adapted output matches the original feature dimension. This padding-and-truncation strategy allows us to flexibly support arbitrary adapter dimensions while maintaining compatibility with the input and output shape requirements of the model. Fig.~\ref{fig:moka} visually summarizes the details of the MoKA approach.

\subsection{Special Case: MoKA$_s$}
A particularly lightweight, and surprisingly effective, variant arises when we fix the right Kronecker factor to the identity matrix, i.e., \( \mathbf{A}_i = \mathbf{I}_{n_{a_i}} \). In this case, the matrix \( \Delta \! \mathbf{W} \) becomes a mixture of learnable block-diagonal matrices. The intuition behind this design is grounded in prior findings~\cite{shi2021sparsebert, pande2020importance}, which show that in the final attention maps of transformer models, \textit{neighboring tokens tend to have higher importance than distant ones}. This leads to
\begin{align}\label{eq11}
\Delta \! \mathbf{W} \mathbf{x} &= \sum_{i=1}^{r} \alpha_i \left( \mathbf{I}_{n_{a_i}} \otimes \mathbf{B}_i \right)\mathbf{x} \\
&= \sum_{i=1}^{r}\alpha_i\mathcal{V}\left\{\mathbf B_i\, \mathcal{R}_{n_{b_i},n_{a_i}}(\mathbf x)\right\},
\end{align}
This block-structured reparameterization allows us to exploit the local importance bias in the attention mechanism while reducing the number of parameters and, \emph{reducing $r$ matrix multiplications at inference}. For additional experimental details and comparisons, see Experiments and Results section.

\section{Theoretical analysis of SGD using MoKA}
\begin{table*}[t]
\centering
\setlength{\tabcolsep}{4pt} 
\renewcommand{\arraystretch}{1.1}
\resizebox{0.97\textwidth}{!}{%
\begin{tabular}{llc|ccccccccc} 
\toprule
\textbf{Model} & \textbf{PEFT Method} &\textbf{\# Trainable Params } &
\textbf{Self-instruct} & \textbf{Longform} & \textbf{Chip2} & \textbf{HH-RLHF} &
\textbf{Unnatural-instruct} & \textbf{Guanaco} & \textbf{Alpaca}  & \textbf{Avg.} \\
\midrule

\multirow{4}{*}{\makecell[l]{LLaMA2-7B}}
\hide{& {\it no fine-tuning}             & 16-bit   &  - & - & - &- & - & - & - & - & 35.10 \\ 
\cmidrule(lr){2-12} }
& QLoRA   &  62.2M & 36.40 & 32.10 & 34.50 & 34.90 & 41.90 & 36.60 & 38.80 & 36.46 \\ 
& QDoRA  & 30.1M  & 38.91 & 35.18 & 38.90 & 42.18 & 43.00 & 41.87 & 40.15 & 40.03 \\ 
& MoKA  & 5.2M  & 41.75 & \textbf{42.90} & \textbf{43.22} & \textbf{42.00} & \textbf{44.11} & \textbf{44.30} & \textbf{43.86} & \textbf{43.16} \\ 
& MoKA$_s$  & \textbf{4.2M}  & \textbf{41.90} & 42.77 & 43.00 & 41.59 & 44.07 & 43.90 & 43.72 & 42.99 \\ 
\midrule
\multirow{4}{*}{\makecell[l]{LLaMA3-8B}}
\hide{& {\it no fine-tuning} & 16-bit   & - & - & - & - & - & -& - & - &  57.38 \\ \cmidrule(lr){2-12} }
& QLoRA           & 56.6M  & 53.46 & 62.58  & 60.07 & 59.20 & 61.00 & 61.15 &   61.69& 59.88 \\
& QDoRA   &  34.4M & 53.85 & 62.60 & 59.22& 59.36 & 60.10 & 62.25 & 61.53 & 59.84 \\ 
& MoKA             & 3.9M & \textbf{56.81} & 63.76 & \textbf{61.48} & 59.81 & \textbf{62.10}& \textbf{64.20} & \textbf{62.69} & \textbf{61.55} \\
& MoKA$_s$           &   \textbf{2.1M} & 56.74 & \textbf{63.90} & 60.64 & \textbf{60.06} & 61.42 & 63.26 & 62.07 & 61.16 \\
\bottomrule
\end{tabular}
}
\caption{Performance comparison of PEFT methods on seven instruction-tuning benchmarks using 4-bit quantized LLaMA2-7B and LLaMA3-8B models, evaluated with 5-shot accuracy on MMLU tasks. Bold numbers indicate the best performance in each block. The 5-shot accuracy of the half-precision LLaMA2-7B and LLaMA3-8B models are 35.10\% and 57.38\%, respectively.}
\label{tab:mmlu}
\end{table*}

In this section, we study the SGD convergence using the Kronecker decomposition of the weight matrix. The training weight matrix is formulated as a shift parameter $\mathbf W + \Delta \! \mathbf W$ where $\mathbf W$ is the pretrained weight matrix and $\Delta \! \mathbf W$ is a trainable matrix that is used to adjust the pretrained matrix $\mathbf W$. 
\hide {For sake of the simplicity in the notation we use $\mathbf W$ and $\Delta \! \mathbf W$ interchangeably as the adapter reparameterization only is a shift transformation in the training parameters and imposes the same training parameter structure to the optimization problem. }

Let's denote $\mathbf U = \Delta \! \mathbf W$ and $\mathbf u = \mathcal{V}(\mathbf U)$, the vectorized weight matrix $mn\times 1$. Let $\mathcal{L}(\mathbf{u})$ be a convex, $L$-smooth loss function. We assume that $\mathbf{U}$ is constrained to be a sum of Kronecker products:

\[
\mathbf{U} = \sum_{i=1}^r \mathbf A_i \otimes \mathbf B_i
\]
with $\mathbf A_i \in \mathbb{R}^{m_{a_i} \times n_{a_i}}$, $\mathbf B_i \in \mathbb{R}^{m_{b_i} \times n_{b_i}}$, such that $m = m_{a_i} m_{b_i}$ and $n = n_{a_i} n_{b_i}$.

We first recall some definitions to confine our attention to differentiable loss function.

\begin{definition}[Convexity]
\label{def:conv}    
A differentiable function \( \mathcal{L}: \mathbb{R}^d \to \mathbb{R} \) is said to be \textbf{convex} if for all \( \mathbf u, \mathbf u' \in \mathbb{R}^d \), the following inequality holds:
\[
\mathcal{L}(\mathbf u') \geq \mathcal{L}( \mathbf u) + \langle \nabla \mathcal{L}( \mathbf u ), \mathbf u' - \mathbf  u \rangle
\]
\end{definition}
This is equivalent to saying that the first-order Taylor approximation of \( \mathcal{L} \) at any point \( \mathbf u \) globally underestimates the function.

\begin{definition}[$L$-Smoothness]
\label{def:lsmooth}
A differentiable function \( \mathcal{L}: \mathbb{R}^d \to \mathbb{R} \) is said to be \textbf{\( L \)-smooth} if $\nabla \mathcal{L}$ is a Lipschitz continuous function, i.e there exists a constant \( L > 0 \) such that for all \( \mathbf  u, \mathbf  u' \in \mathbb{R}^d \), the gradient satisfies
\[
\| \nabla \mathcal{L}(\mathbf u) - \nabla \mathcal{L}(\mathbf u') \| \leq L \| \mathbf u - \mathbf u' \|.
\]
Equivalently, for all \( \mathbf  u, \mathbf u' \in \mathbb{R}^d \), the function satisfies the upper bound
\[
\mathcal{L}(\mathbf u') \leq \mathcal{L}(\mathbf u) + \langle \nabla \mathcal{L}(\mathbf u), \mathbf u' - \mathbf u \rangle + \frac{L}{2} \| \mathbf u' - \mathbf u \|^2.
\]
\end{definition}
Before presenting the main result of this section, we briefly recall the SGD update rule and state the assumptions required for our proof. At iteration \(t\), the weights are updated as  
\[
\mathbf{u}_{t+1} = \mathbf{u}_t - \eta\,\nabla \mathcal{L}(\mathbf{u}_t; \xi_t), 
\quad t = 0,1,\dotsc,T-1,
\]
where \(\eta > 0\) is the learning rate, and \(\nabla \mathcal{L}(\mathbf{u}_t; \xi_t)\) denotes the estimated gradient evaluated on a mini-batch sampled using the random seed \(\xi_t\).  
Throughout the analysis, we assume that the seeds \(\{\xi_t\}_{t=0}^{T-1}\) are i.i.d. random variables.

We break the convergence into two parts 
$$\mathcal L(\mathbf u) - \mathcal L^* = \underbrace{\mathcal L(\mathbf u) - \mathcal{L}_{\min}}_{\text{optimization error}} +\underbrace{\mathcal{L}_{\min} - \mathcal L^*}_{\text{approximation error}},$$
where $\mathcal L^*$ is the minimum loss in the unconstrained space, and $\mathcal{L}_{\min}$ is the minimum in the Kronecker-structured space. The approximation error will remain there and vanishes if the minimum loss falls in the Kronecker-structured space. We study the \emph{optimization error} through the Frobenius norm of the gradients that clarifies how fast $\mathcal L(\mathbf u)$ approaches to the minimum in the Kronecker structured space $\mathcal{L}_{\min}$.
Without loss of generality, we rescale  Kronecker weights $\sqrt{\alpha_i}\mathbf A_i, \sqrt{\alpha_i}\mathbf B_i$ that plays no role in our theoretical study.  
To study the convergence rate of the gradient we need the following assumptions: 
\begin{enumerate}[label=\textbf{A\arabic*)}, leftmargin=*, labelsep=0.5em]

  \item $\mathcal{L}$ is an $L$-smooth convex function.
    \item $\mathcal{L}(\mathbf{u}_t)$ is bounded and  the estimated gradients are unbiased estimator of
the true gradients of the loss function, i.e.
    \[
    \mathbb{E}_{\xi_t}[\nabla \mathcal{L}(\mathbf{u}_t; {\xi_t}) ] = \nabla \mathcal{L}(\mathbf{u}_t), \,\,\, \forall \mathbf{u}
    \] 
    \item Kronecker factors and their gradient are bounded, i.e.
    \resizebox{\linewidth}{!}{
\(
\mathbb{E}_{\xi_t} \left[\left( \| \nabla_{\mathbf{A}_i} \mathcal{L}(\mathbf{u}_t; \xi_t) \|_F^2 \| \mathbf{B}_i \|_F^2 + \| \mathbf{A}_i \|_F^2 \| \nabla_{\mathbf{B}_i} \mathcal{L}(\mathbf{u}_t; \xi_t) \|_F^2 \right) \right] \leq G,
\)
}

    For all $\mathbf{u}$ and $i=1,2, \cdots, r$, where $\nabla_{\mathbf{A}_i} \mathcal{L}$ and $\nabla_{\mathbf{B}_i} \mathcal{L}$ are parts of the gradients that correspond to elements of $\mathbf A_i$ and $\mathbf B_i$, respectively. Also, $G$ is a positive constant.
\end{enumerate}

\begin{theorem}
\label{theo:kron}    
Suppose assumptions A1, A2, and A3 hold. Then  
\[
\frac{1}{T} \sum_{t=0}^{T-1} \mathbb{E}_{\xi_t} \left[ \| \nabla \mathcal{L}(\mathbf{u}_t; {\xi_t}) \|_F^2 \right] \leq \frac{\mathcal{L}(\mathbf{u}_0) - \mathcal{L}_{\min}}{\eta T} + \frac{\eta L G}{2},
\]
 Where $\mathcal{L}_{\min}$ is the minimum of the loss within the class of matrices representable by the sum of $r$ Kronecker products. 




\end{theorem}
See Appendix for the proof. Theorem 1 essentially shows that the average gradients through the steps of SGD with a constant learning rate $\eta$ is composed of two terms, the first term is $\mathcal O(\frac{1}{T})$, and the second term becomes negligible for a small rate $\eta$. This is the same behavior as the unconstrained case. This upper bound is minimized for the choice of $\eta^* = \sqrt{\frac{2(\mathcal{L}(\mathbf{u}_0) - \mathcal{L}_{\min})}{LGT}}$ which makes the convergence of both terms of order $\mathcal O(\frac{1}{\sqrt{T}})$. See Appendix for the details. 

\section{Experiments and Results} \label{experiments}

One important function of adapters is to mitigate the accuracy degradation caused by low-bit quantization of LLMs. In this section, we demonstrate how our proposed method, MoKA, effectively addresses this challenge by restoring performance to low-bit quantized versions of LLaMA2-7B and LLaMA3-8B models, following the quantization settings detailed in~\cite{qlora}. First, we outline the experimental settings of MoKA. Then, we evaluate MoKA against two competitive and widely adopted baselines: QLoRA~\cite{qlora} and QDoRA~\cite{dora}.

We conduct our evaluation under two distinct experimental scenarios. In the first scenario, we adopt an instruction-tuning setup as defined in~\cite{qlora}. This scenario is particularly challenging for low-rank methods, since task-specific knowledge often requires higher-dimensional parameterization. In the second scenario, we benchmark our approach using commonsense reasoning tasks, a well-established evaluation framework frequently employed to compare the effectiveness of different adapter techniques.




\begin{table*}[t]
\centering
\setlength{\tabcolsep}{4pt} 
\renewcommand{\arraystretch}{1.1}
\resizebox{0.97\textwidth}{!}{%
\begin{tabular}{lllc|ccccccccc} 
\toprule
\textbf{Model} & \textbf{PEFT Method}& \textbf{Model bits} & \textbf{\# Trainable Params} &
\textbf{BoolQ} & \textbf{PIQA} & \textbf{SIQA} & \textbf{HellaSwag} &
\textbf{WinoGrande} & \textbf{ARC-e} & \textbf{ARC-c} & \textbf{OBQA} & \textbf{Avg.} \\
\midrule

\multirow{5}{*}{\makecell[l]{LLaMA2-7B}}
& no fine-tuning       & 16-bit       &  - & 71.10 & 78.34 & 43.90 & 56.70 & 68.08 & 69.27 & 39.93 & 31.80 & 57.39 \\ \cmidrule(lr){2-13} 
& QLoRA               &4-bit    & 62.2M & 73.86 & 77.61 & 47.90 & 54.13 & \textbf{71.03} & 70.37 & 40.69 & 31.20 & 58.35 \\
& QDoRA  & 4-bit &30.1M & 74.94 & 75.78 & \textbf{48.46} & 55.37& 69.00 & 70.49 & 40.25 & 32.00 &  58.29 \\
& MoKA            & 4-bit &5.2M  & \textbf{75.74} & 78.12 & 48.25 & \textbf{56.14} & 70.00 & \textbf{72.76} & \textbf{41.21} & \textbf{32.80}& \textbf{59.38} \\
& MoKA$_s$           & 4-bit &\textbf{4.2M}  & 75.13 & \textbf{78.18} & 46.46 & 55.70 & 68.11 & 71.29 & 40.52 & 32.00 & 58.42 \\
\midrule
\multirow{5}{*}{\makecell[l]{LLaMA3-8B}}
& no fine-tuning            & 16-bit &- & 82.59 & 77.74 & 49.43 & 57.04 & 71.74 & 80.09 & 52.73 & 32.90 & 60.24 \\ \cmidrule(lr){2-13} 
& QLoRA                   & 4-bit &56.6M & 84.33 & 78.61 & 45.85 & 58.11 & \textbf{75.45} & 79.71 & 48.80 & 32.40 & 62.91 \\
& QDoRA& 4-bit &34.4M & 84.11 & 78.83 & 46.05 & 57.69 & 75.13 & 79.12 & 48.37 & 32.20 &  62.69  \\
& MoKA            & 4-bit &3.9M & 83.82 & 79.27 & 46.57 & \textbf{58.79} & 73.71 & \textbf{81.06} & \textbf{50.30} & 33.00 & 63.29 \\
& MoKA$_s$            & 4-bit &\textbf{2.1M} & \textbf{84.72} & \textbf{79.78} & \textbf{47.08} & 58.29 & 75.37 & 80.63 & 49.74 & \textbf{33.20} & \textbf{63.60} \\
\bottomrule
\end{tabular}
}
\caption{Performance comparison of PEFT methods on commonsense reasoning datasets using 4-bit quantized LLaMA2-7B and LLaMA3-8B models. Bold numbers denote the best performance in each block. During fine-tuning, the training sets of all datasets are concatenated. For evaluation, zero-shot accuracy is reported on each test set individually.
}
\label{tab:peft-results}
\end{table*}

\begin{table*}[t]
\centering
\setlength{\tabcolsep}{4pt} 
\renewcommand{\arraystretch}{1.1}
\resizebox{0.8\textwidth}{!}{%
\begin{tabular}{llcccccccccc} 
\toprule
\textbf{Model} & \textbf{PEFT Method} & 
\textbf{BoolQ} & \textbf{PIQA} & \textbf{SIQA} & \textbf{HellaSwag} &
\textbf{WinoGrande} & \textbf{ARC-e} & \textbf{ARC-c} & \textbf{OBQA} & \textbf{Avg.} \\
\midrule

\multirow{4}{*}{\makecell[l]{LLaMA2-7B}}
& MoKA   w/o gates          & 74.49 & 77.28 & 47.23 & 55.80 & 69.13 & 71.92 & 41.02 & 32.00 & 58.61 \\
& MoKA             & \textbf{75.74} & \textbf{78.12} & \textbf{48.25} & \textbf{56.14} & \textbf{70.00} & \textbf{72.76} & \textbf{41.21} & \textbf{32.80}& \textbf{59.38} \\ \cmidrule(lr){2-11}
& MoKA$_s$  w/o gates           & 74.32 & 77.89 & 46.00 & \textbf{56.10} & 67.95 & 71.00 & 40.36 & 31.80 & 58.18 \\

& MoKA$_s$             & \textbf{75.13} & \textbf{78.18} & \textbf{46.46} & 55.70 & \textbf{68.11} & \textbf{71.29} & \textbf{40.52} & \textbf{32.00} & \textbf{58.42} \\
\midrule
\multirow{4}{*}{\makecell[l]{LLaMA3-8B}}
& MoKA  w/o gates           & 80.36 & 79.21 & 46.10 & 56.81 & 73.60 & 80.27 & 49.74 & \textbf{33.40} & 62.44 \\
& MoKA           & \textbf{83.82} & \textbf{79.27} & \textbf{46.57} & \textbf{58.79} & \textbf{73.71} & \textbf{81.06} & \textbf{50.30} & 33.00 & \textbf{63.29} \\
\cmidrule(lr){2-11}
& MoKA$_s$  w/o gates           & 82.56 & 78.89 & 46.33 & 58.13 &74.58  & 80.10 & 49.50 & 32.60 & 62.84 \\
& MoKA$_s$            & \textbf{84.72} & \textbf{79.78} & \textbf{47.08} & \textbf{58.29} & \textbf{75.37} & \textbf{80.63} & \textbf{49.74} & \textbf{33.20} & \textbf{63.60} \\
\bottomrule
\end{tabular}
}
\caption{Impact of the gating mechanism on MoKA and MoKA$_s$ across commonsense reasoning datasets. Bold numbers indicate the best performance in each block.}
\label{tab:gating}
\end{table*}

\subsection{MoKA Experimental Settings}
As a common practice in fine-tuning LLMs~\cite{lora}, we attach MoKA adapters to the \textit{query} ($q$) and \textit{value} ($v$) projection matrices of every transformer layer. In addition, we use effective batch size of $32$, and AdamW optimizer~\cite{adamw} with the learning rate of $2\times 10^{-4}$.

\paragraph{LLaMA2-7B:}  
Each projection receives ten Kronecker components composed of the following five filter shapes, each instantiated twice:
\begin{align*}
(\mathbf{A}_i, \mathbf{B}_i) \in \Big\{
&(64{\times}64,\;64{\times}64),\quad
(32{\times}128,\;128{\times}32),\\
&(128{\times}32,\;32{\times}128),\quad
(16{\times}256,\;256{\times}16),\\
&(256{\times}16,\;16{\times}256)
\Big\}.
\end{align*}

\paragraph{LLaMA3-8B:}  
The query projection uses the same filter shapes as in LLaMA2-7B. For the value projection, we use:
\begin{align*}
(\mathbf{A}_i, \mathbf{B}_i) \in \Big\{
&(32{\times}64,\;32{\times}64),\quad
(16{\times}128,\;64{\times}32),\\
&(64{\times}32,\;16{\times}128),\quad
(8{\times}256,\;128{\times}16),\\
&(128{\times}16,\;8{\times}256)
\Big\},
\end{align*}
again instantiated twice per shape.

\paragraph{MoKA$_s$:}  
Each \( \mathbf{A}_i \) is replaced by the identity matrix, and for \( \mathbf{B}_i \in \mathbb{R}^{p \times p} \), \( p \) is a prime number in the range 2–97. For value projection in LLaMA3-8B, we use filters of shape \( p \times 4p \) to match the layer dimensions. All components are combined using a learnable softmax gating mechanism, as described in Methodology Section, yielding a flexible and hardware-friendly adapter.

\subsection{Instruction-tuning Experiments}
In this section, we present comprehensive experiments focused on instruction-tuning. To provide an in-depth exploration of this fine-tuning approach, we follow a widely accepted experimental framework. Specifically, we fine-tune 4-bit quantized versions of the LLaMA2-7B and LLaMA3-8B models across seven instruction-tuning datasets: Self-Instruct~\cite{wang2022self}, Longform~\cite{koksal2023longform}, Chip2~\cite{ma2024investigating}, HH-RLHF~\cite{bai2022training}, Unnatural-Instruct~\cite{honovich2022unnatural}, Guanaco~\cite{kopf2023openassistant}, and Alpaca~\cite{taori2023stanford}. To assess instruction-tuning performance, we report 5-shot accuracy on the Massively Multitask Language Understanding (MMLU) benchmark~\cite{hendrycks2020measuring}, summarized in Table~\ref{tab:mmlu}.
The key observations from this table are: 

\begin{enumerate}
    \item MoKA achieves an average improvement of 8.6\% over the half-precision baseline on LLaMA2-7B and 4.17\% on LLaMA3-8B, highlighting the effectiveness of our training methodology.    
    \item Compared to QLoRA, MoKA delivers a 6.7\% average improvement on LLaMA2-7B with approximately 12$\times$ fewer trainable parameters. Similarly, for LLaMA3-8B, MoKA offers a 1.67\% average gain, using roughly 14$\times$ fewer parameters.
    \item Against the QDoRA baseline, MoKA improves performance by 3.13\% on LLaMA2-7B, with about 6$\times$ fewer parameters. For LLaMA3-8B, it surpasses QDoRA by 1.71\%, while reducing the number of trainable parameters from 34.4M to just 3.9M, around 9$\times$ fewer parameters.
    \item MoKA$_s$ maintains performance comparable to MoKA while further reducing the parameter count: 2.1M for LLaMA2-7B and 4.2M for LLaMA3-8B. This result shows efficiency and scalability of MoKA$_s$ for highly efficient low-bit adaptation.
\end{enumerate}

We attribute these improvements to two primary factors. First, MoKA introduces multiple filter shapes, providing increased flexibility by removing the low-rank constraint and structural assumptions present in other methods. Second, the use of a gating mechanism allows MoKA to prioritize the most informative components during adaptation. Moreover, the compact variant MoKA$_s$ attains comparable performance with even fewer parameters by exploiting the local attention bias of transformer architectures~\cite{shi2021sparsebert, pande2020importance}.

\subsection{Commonsense Reasoning Experiments}
In Table~\ref{tab:peft-results}, we evaluate our proposed MoKA method against QLoRA and QDoRA, using both 4-bit quantized LLaMA2-7B and LLaMA3-8B models, on a suite of zero-shot commonsense reasoning tasks. For reference, we also report the accuracy of the original half-precision models without fine-tuning or quantization, denoted as \textit{no fine-tuning}.

The commonsense reasoning benchmark~\cite{hu-etal-2023-llm} comprises eight sub-tasks, each with standard training and testing splits. Following the protocol established in~\cite{hu-etal-2023-llm}, we merge the training sets of all eight tasks into a single joint training set, and evaluate performance individually on the test set of each task.

From Table~\ref{tab:peft-results}, we observe that MoKA not only improves over its quantized counterparts but also surpasses the half-precision baseline in average accuracy. MoKA achieves nearly a 2\% gain on LLaMA2-7B and more than 3\% on LLaMA3-8B, despite operating with 4-bit quantized pretrained weights.
Furthermore, MoKA outperforms QLoRA by more than 1\% on average for LLaMA2-7B and approximately 0.4\% for LLaMA3-8B, while using over 12$\times$ fewer trainable parameters. Compared to QDoRA, MoKA achieves gains of over 1\% and around 0.6\% higher accuracy on LLaMA2-7B and LLaMA3-8B, respectively, with reductions of approximately 6$\times$ and 9$\times$ in trainable parameters.
We attribute these improvements to MoKA’s greater expressive capacity, enabled by its gated mixture of Kronecker adapters, which overcomes the representational limitations  imposed by the low-rank constraints in QLoRA and QDoRA.

Similar to instruction-tuning experiments, MoKA$_s$ achieves comparable performance to MoKA while using even fewer trainable parameters. This efficiency arises from its ability to exploit local token-level importance patterns in the attention mechanism.

\subsection{Effectiveness of gating mechanism}
In Table~\ref{tab:gating}, we study the effect of our proposed gating mechanism by comparing MoKA and MoKA$_s$ with their respective ungated variants, denoted as MoKA w/o gates and MoKA$_s$ w/o gates. In the ungated setting, the outputs of all adapters are averaged uniformly (i.e., $\alpha_i = 1/r$, where $r$ is the number of adapters), removing the adaptive weighting learned during training. This simplification allows us to isolate the contribution of the gating mechanism itself.

As shown in the table, the inclusion of gating consistently improves performance across all models. For LLaMA2-7B, MoKA and MoKA$_s$ outperform their ungated counterparts by 0.77 and 0.24 points, respectively, yielding an average improvement of 0.51. For LLaMA3-8B, the gains are more pronounced: MoKA improves by 0.85 and MoKA$_s$ by 0.76, averaging 0.81. These results demonstrate that the gating mechanism enables more effective adapter integration by learning to emphasize informative components rather than treating all adapters equally.

This confirms our hypothesis that different adapters contribute unequally depending on the input and task, and that a learnable gating function can adaptively prioritize more relevant features. The improved performance across multiple commonsense reasoning tasks suggests that gating is a crucial component in achieving robust parameter-efficient fine-tuning.

\vspace{-0.8em}
\section{Conclusion}

This work introduces MoKA, a novel parameter-efficient fine-tuning approach that overcomes the representational bottlenecks of traditional low-rank approaches by employing a gated mixture of Kronecker adapters. Unlike existing PEFT methods, MoKA offers greater expressiveness through rank flexibility and dynamic weighting of multiple Kronecker components. To support efficient deployment, we reformulate Kronecker operations using standard matrix multiplications, ensuring compatibility with GPU-accelerated environments.

Our experiments on instruction-tuning and commonsense reasoning tasks, conducted on 4-bit quantized LLaMA2-7B and LLaMA3-8B models, demonstrate that MoKA consistently outperforms strong PEFT baselines such as QLoRA and QDoRA, while drastically reducing the number of trainable parameters. These results establish MoKA as a compelling solution for adapting large language models in resource-constrained settings, combining high performance with low computational overhead.

Future work may explore extending MoKA to other architectures, integrating it with more advanced quantization techniques, or applying its principles to other modalities beyond language.

\bibliography{aaai2026}

\begin{thebibliography}{29}
\providecommand{\natexlab}[1]{#1}

\bibitem[{Bai et~al.(2022)Bai, Jones, Ndousse, Askell, Chen, DasSarma, Drain, Fort, Ganguli, Henighan et~al.}]{bai2022training}
Bai, Y.; Jones, A.; Ndousse, K.; Askell, A.; Chen, A.; DasSarma, N.; Drain, D.; Fort, S.; Ganguli, D.; Henighan, T.; et~al. 2022.
\newblock Training a helpful and harmless assistant with reinforcement learning from human feedback.
\newblock \emph{arXiv preprint arXiv:2204.05862}.

\bibitem[{Ben~Zaken, Goldberg, and Ravfogel(2022)}]{ben-zaken-etal-2022-bitfit}
Ben~Zaken, E.; Goldberg, Y.; and Ravfogel, S. 2022.
\newblock {B}it{F}it: Simple Parameter-efficient Fine-tuning for Transformer-based Masked Language-models.
\newblock In Muresan, S.; Nakov, P.; and Villavicencio, A., eds., \emph{Proceedings of the 60th Annual Meeting of the Association for Computational Linguistics (Volume 2: Short Papers)}, 1--9. Dublin, Ireland: Association for Computational Linguistics.

\bibitem[{Braga et~al.(2024)Braga, Raganato, Pasi et~al.}]{braga2024adakron}
Braga, M.; Raganato, A.; Pasi, G.; et~al. 2024.
\newblock Adakron: An adapter-based parameter efficient model tuning with kronecker product.
\newblock In \emph{2024 Joint International Conference on Computational Linguistics, Language Resources and Evaluation, LREC-COLING 2024-Main Conference Proceedings}, 350--357.

\bibitem[{Dettmers et~al.(2023)Dettmers, Pagnoni, Holtzman, and Zettlemoyer}]{qlora}
Dettmers, T.; Pagnoni, A.; Holtzman, A.; and Zettlemoyer, L. 2023.
\newblock Qlora: Efficient finetuning of quantized llms.
\newblock \emph{Advances in neural information processing systems}, 36: 10088--10115.

\bibitem[{Edalati et~al.(2022)Edalati, Tahaei, Kobyzev, Nia, Clark, and Rezagholizadeh}]{Krona}
Edalati, A.; Tahaei, M.; Kobyzev, I.; Nia, V.~P.; Clark, J.~J.; and Rezagholizadeh, M. 2022.
\newblock KronA: Parameter Efficient Tuning with Kronecker Adapter.
\newblock arXiv:2212.10650.

\bibitem[{Hendrycks et~al.(2020)Hendrycks, Burns, Basart, Zou, Mazeika, Song, and Steinhardt}]{hendrycks2020measuring}
Hendrycks, D.; Burns, C.; Basart, S.; Zou, A.; Mazeika, M.; Song, D.; and Steinhardt, J. 2020.
\newblock Measuring massive multitask language understanding.
\newblock \emph{arXiv preprint arXiv:2009.03300}.

\bibitem[{Honovich et~al.(2022)Honovich, Scialom, Levy, and Schick}]{honovich2022unnatural}
Honovich, O.; Scialom, T.; Levy, O.; and Schick, T. 2022.
\newblock Unnatural instructions: Tuning language models with (almost) no human labor.
\newblock \emph{arXiv preprint arXiv:2212.09689}.

\bibitem[{Houlsby et~al.(2019)Houlsby, Giurgiu, Jastrzebski, Morrone, De~Laroussilhe, Gesmundo, Attariyan, and Gelly}]{houlsby2019parameter}
Houlsby, N.; Giurgiu, A.; Jastrzebski, S.; Morrone, B.; De~Laroussilhe, Q.; Gesmundo, A.; Attariyan, M.; and Gelly, S. 2019.
\newblock Parameter-Efficient Transfer Learning for {NLP}.
\newblock In Chaudhuri, K.; and Salakhutdinov, R., eds., \emph{Proceedings of the 36th International Conference on Machine Learning}, volume~97 of \emph{Proceedings of Machine Learning Research}, 2790--2799. PMLR.

\bibitem[{Hu et~al.(2022)Hu, Shen, Wallis, Allen-Zhu, Li, Wang, Wang, Chen et~al.}]{lora}
Hu, E.~J.; Shen, Y.; Wallis, P.; Allen-Zhu, Z.; Li, Y.; Wang, S.; Wang, L.; Chen, W.; et~al. 2022.
\newblock Lora: Low-rank adaptation of large language models.
\newblock \emph{ICLR}, 1(2): 3.

\bibitem[{Hu et~al.(2023)Hu, Wang, Lan, Xu, Lim, Bing, Xu, Poria, and Lee}]{hu-etal-2023-llm}
Hu, Z.; Wang, L.; Lan, Y.; Xu, W.; Lim, E.-P.; Bing, L.; Xu, X.; Poria, S.; and Lee, R. 2023.
\newblock {LLM}-Adapters: An Adapter Family for Parameter-Efficient Fine-Tuning of Large Language Models.
\newblock In Bouamor, H.; Pino, J.; and Bali, K., eds., \emph{Proceedings of the 2023 Conference on Empirical Methods in Natural Language Processing}, 5254--5276. Singapore: Association for Computational Linguistics.

\bibitem[{K{\"o}ksal et~al.(2023)K{\"o}ksal, Schick, Korhonen, and Sch{\"u}tze}]{koksal2023longform}
K{\"o}ksal, A.; Schick, T.; Korhonen, A.; and Sch{\"u}tze, H. 2023.
\newblock Longform: Optimizing instruction tuning for long text generation with corpus extraction.
\newblock \emph{arXiv preprint arXiv:2304.08460}.

\bibitem[{K{\"o}pf et~al.(2023)K{\"o}pf, Kilcher, Von~R{\"u}tte, Anagnostidis, Tam, Stevens, Barhoum, Nguyen, Stanley, Nagyfi et~al.}]{kopf2023openassistant}
K{\"o}pf, A.; Kilcher, Y.; Von~R{\"u}tte, D.; Anagnostidis, S.; Tam, Z.~R.; Stevens, K.; Barhoum, A.; Nguyen, D.; Stanley, O.; Nagyfi, R.; et~al. 2023.
\newblock Openassistant conversations-democratizing large language model alignment.
\newblock \emph{Advances in neural information processing systems}, 36: 47669--47681.

\bibitem[{Lester, Al-Rfou, and Constant(2021)}]{lester-etal-2021-power}
Lester, B.; Al-Rfou, R.; and Constant, N. 2021.
\newblock The Power of Scale for Parameter-Efficient Prompt Tuning.
\newblock In Moens, M.-F.; Huang, X.; Specia, L.; and Yih, S. W.-t., eds., \emph{Proceedings of the 2021 Conference on Empirical Methods in Natural Language Processing}, 3045--3059. Online and Punta Cana, Dominican Republic: Association for Computational Linguistics.

\bibitem[{Li and Liang(2021)}]{li2021prefix}
Li, X.~L.; and Liang, P. 2021.
\newblock Prefix-tuning: Optimizing continuous prompts for generation.
\newblock \emph{arXiv preprint arXiv:2101.00190}.

\bibitem[{Liu et~al.(2024)Liu, Wang, Yin, Molchanov, Wang, Cheng, and Chen}]{dora}
Liu, S.-Y.; Wang, C.-Y.; Yin, H.; Molchanov, P.; Wang, Y.-C.~F.; Cheng, K.-T.; and Chen, M.-H. 2024.
\newblock Dora: Weight-decomposed low-rank adaptation.
\newblock In \emph{Forty-first International Conference on Machine Learning}.

\bibitem[{Loshchilov and Hutter(2019)}]{adamw}
Loshchilov, I.; and Hutter, F. 2019.
\newblock Decoupled Weight Decay Regularization.
\newblock arXiv:1711.05101.

\bibitem[{Ma, Li, and Shang(2024)}]{ma2024investigating}
Ma, R.; Li, W.; and Shang, F. 2024.
\newblock Investigating Public Fine-Tuning Datasets: A Complex Review of Current Practices from a Construction Perspective.
\newblock \emph{arXiv preprint arXiv:2407.08475}.

\bibitem[{Mao et~al.(2022)Mao, Mathias, Hou, Almahairi, Ma, Han, Yih, and Khabsa}]{mao-etal-2022-unipelt}
Mao, Y.; Mathias, L.; Hou, R.; Almahairi, A.; Ma, H.; Han, J.; Yih, W.-t.; and Khabsa, M. 2022.
\newblock UniPELT: A Unified Framework for Parameter-Efficient Language Model Tuning.
\newblock In \emph{Proceedings of the 60th Annual Meeting of the Association for Computational Linguistics}.

\bibitem[{Pande et~al.(2020)Pande, Budhraja, Nema, Kumar, and Khapra}]{pande2020importance}
Pande, M.; Budhraja, A.; Nema, P.; Kumar, P.; and Khapra, M.~M. 2020.
\newblock On the importance of local information in transformer based models.
\newblock \emph{arXiv preprint arXiv:2008.05828}.

\bibitem[{Rajabzadeh et~al.(2024)Rajabzadeh, Valipour, Zhu, Tahaei, Kwon, Ghodsi, Chen, and Rezagholizadeh}]{rajabzadeh-etal-2024-qdylora}
Rajabzadeh, H.; Valipour, M.; Zhu, T.; Tahaei, M.~S.; Kwon, H.~J.; Ghodsi, A.; Chen, B.; and Rezagholizadeh, M. 2024.
\newblock {QD}y{L}o{RA}: Quantized Dynamic Low-Rank Adaptation for Efficient Large Language Model Tuning.
\newblock In Dernoncourt, F.; Preo{\c{t}}iuc-Pietro, D.; and Shimorina, A., eds., \emph{Proceedings of the 2024 Conference on Empirical Methods in Natural Language Processing: Industry Track}, 712--718. Miami, Florida, US: Association for Computational Linguistics.

\bibitem[{Shi et~al.(2021)Shi, Gao, Ren, Xu, Liang, Li, and Kwok}]{shi2021sparsebert}
Shi, H.; Gao, J.; Ren, X.; Xu, H.; Liang, X.; Li, Z.; and Kwok, J. T.-Y. 2021.
\newblock Sparsebert: Rethinking the importance analysis in self-attention.
\newblock In \emph{International Conference on Machine Learning}, 9547--9557. PMLR.

\bibitem[{Shi et~al.(2024)Shi, Yao, Li, Zhang, and Zhao}]{shi2024reference}
Shi, L.; Yao, Y.; Li, Z.; Zhang, L.; and Zhao, H. 2024.
\newblock Reference Trustable Decoding: A Training-Free Augmentation Paradigm for Large Language Models.
\newblock \emph{Advances in Neural Information Processing Systems}, 37: 80034--80055.

\bibitem[{Taori et~al.(2023)Taori, Gulrajani, Zhang, Dubois, Li, Guestrin, Liang, and Hashimoto}]{taori2023stanford}
Taori, R.; Gulrajani, I.; Zhang, T.; Dubois, Y.; Li, X.; Guestrin, C.; Liang, P.; and Hashimoto, T.~B. 2023.
\newblock Stanford alpaca: An instruction-following llama model.

\bibitem[{Valipour et~al.(2022)Valipour, Rezagholizadeh, Kobyzev, and Ghodsi}]{valipour2022dylora}
Valipour, M.; Rezagholizadeh, M.; Kobyzev, I.; and Ghodsi, A. 2022.
\newblock Dylora: Parameter efficient tuning of pre-trained models using dynamic search-free low-rank adaptation.
\newblock \emph{arXiv preprint arXiv:2210.07558}.

\bibitem[{Van~Loan and Pitsianis(1993)}]{VanLoanPitsianis_KronApprox_1993}
Van~Loan, C.~F.; and Pitsianis, N. 1993.
\newblock Approximation with Kronecker products.
\newblock In \emph{Linear algebra for large scale and real-time applications}, 293--314. Springer.

\bibitem[{Wang et~al.(2025)Wang, Chen, Jiang, Pan, Cai, Yang, and Yang}]{Wang2025-PEFTsurvey}
Wang, L.; Chen, S.; Jiang, L.; Pan, S.; Cai, R.; Yang, S.; and Yang, F. 2025.
\newblock Parameter-efficient fine-tuning in large language models: a survey of methodologies.
\newblock \emph{Artificial Intelligence Review}, 58(8): 227.

\bibitem[{Wang et~al.(2022)Wang, Kordi, Mishra, Liu, Smith, Khashabi, and Hajishirzi}]{wang2022self}
Wang, Y.; Kordi, Y.; Mishra, S.; Liu, A.; Smith, N.~A.; Khashabi, D.; and Hajishirzi, H. 2022.
\newblock Self-instruct: Aligning language models with self-generated instructions.
\newblock \emph{arXiv preprint arXiv:2212.10560}.

\bibitem[{Zhang et~al.(2023)Zhang, Chen, Bukharin, He, Cheng, Chen, and Zhao}]{adalora}
Zhang, Q.; Chen, M.; Bukharin, A.; He, P.; Cheng, Y.; Chen, W.; and Zhao, T. 2023.
\newblock Adaptive Budget Allocation for Parameter-Efficient Fine-Tuning.
\newblock In \emph{The Eleventh International Conference on Learning Representations}.

\bibitem[{Zhou et~al.(2024)Zhou, Wan, Vuli{\'c}, and Korhonen}]{zhou-etal-2024-autopeft}
Zhou, H.; Wan, X.; Vuli{\'c}, I.; and Korhonen, A. 2024.
\newblock {A}uto{PEFT}: Automatic Configuration Search for Parameter-Efficient Fine-Tuning.
\newblock \emph{Transactions of the Association for Computational Linguistics}, 12: 525--542.

\end{thebibliography}

\section*{Appendix}
\subsection*{Proof of Theorem~1:}
At iteration \(t\), let $\mathbf{U}_t = \sum_{i=1}^r \mathbf{A}_i^{(t)} \otimes \mathbf{B}_i^{(t)}$ and $\mathbf u_t = \mathcal{V}(\mathbf U_t)$, the vectorized representation of $\mathbf U_t$. Due to $L$-smoothness of $\mathcal{L}$:
\[
\mathcal{L}(\mathbf{u}_{t+1}) \leq \mathcal{L}(\mathbf{u}_t) + \langle \nabla \mathcal{L}(\mathbf{u}_t), \mathbf{u}_{t+1} - \mathbf{u}_t \rangle + \frac{L}{2} \| \mathbf{u}_{t+1} - \mathbf{u}_t \|_F^2 \ .
\]
The weights are updated over one step as
\[
\mathbf{u}_{t+1} = \mathbf{u}_t - \eta\,\nabla \mathcal{L}(\mathbf{u}_t; \xi_t),
\]
where \(\nabla \mathcal{L}(\mathbf{u}_t; \xi_t)\) denotes the estimated gradient evaluated on a mini-batch sampled using the random seed \(\xi_t\). Hereafter, we use \(\nabla\mathcal{L}_t\) and \(\nabla \mathcal{L}(\mathbf{u}_t; \xi_t)\) interchangeably for the sake of simplicity in writing.  

The weight update over one step can be re-written as
\[
\mathbf{u}_{t+1} = \sum_{i=1}^r \mathcal{V}\left\{( \mathbf{A}_i^{(t)}\!-\!\eta \nabla_{\mathbf{A}_i} \mathcal{L}_t) \otimes ( \mathbf{B}_i^{(t)}\!-\!\eta \nabla_{\mathbf{B}_i} \mathcal{L}_t )\right\},
\]
where $\nabla_{\mathbf{A}_i} \mathcal{L}_t$ and $\nabla_{\mathbf{B}_i} \mathcal{L}_t$ are parts of the gradients that correspond to elements of $\mathbf A_i$ and $\mathbf B_i$, respectively.

Linearizing the equation and collecting terms, the leading-order term is
\[
\mathbf{u}_{t+1} - \mathbf{u}_t \approx -\eta \underbrace{\sum_{i=1}^r \mathcal{V}\left\{ \nabla_{\mathbf{A}_i} \mathcal{L}_t \otimes \mathbf{B}_i^{(t)} + \mathbf{A}_i^{(t)} \otimes \nabla_{\mathbf{B}_i} \mathcal{L}_t \right\}}_{\mathcal{G}_t}.
\]
Using the leading-order term of \(\mathbf{u}_{t+1} - \mathbf{u}_t\), we re-write the $L$-smoothness inequality of $\mathcal{L}$ after taking expectation over random seed \(\xi_t\), such that 
\begin{align*}
\mathbb{E}_{\xi_t}[ \mathcal{L}(\mathbf{u}_{t+1}; \xi_t) ] \leq \mathbb{E}_{\xi_t}&[ \mathcal{L}(\mathbf{u}_t; \xi_t) ] - \eta \mathbb{E}_{\xi_t}\! \left[ \left\langle \nabla \mathcal{L}(\mathbf{u}_t; \xi_t), \mathcal{G}_t \right\rangle \right] \\&+ \frac{L \eta^2}{2} \mathbb{E}_{\xi_t} \! \left[ \left\| \mathcal{G}_t \right\|_F^2 \right].
\end{align*}
Considering that \(\mathcal{G}_t\) is an approximation of \(\nabla \mathcal{L}(\mathbf{u}_t; \xi_t)\), we can replace
\[
\mathbb{E}_{\xi_t}\! \left[ \left\langle \nabla \mathcal{L}(\mathbf{u}_t; \xi_t), \mathcal{G}_t \right\rangle \right]
\]
with
\[
\mathbb{E}_{\xi_t} \left[ \| \nabla \mathcal{L}(\mathbf{u}_t;\xi_t) \|_F^2 \right].
\]
In addition, by expanding \(\left\| \mathcal{G}_t \right\|_F^2\), we get
\begin{align*}
\left\| \mathcal{G}_t \right\|_F^2 &= \left\| \sum_{i=1}^r \mathcal{V}\left\{ \nabla_{\mathbf{A}_i} \mathcal{L}_t \!\otimes\! \mathbf{B}_i^{(t)} + \mathbf{A}_i^{(t)} \!\otimes\! \nabla_{\mathbf{B}_i} \mathcal{L}_t \right\} \right\|_F^2\\
&\underset{(1)}{\leq} \sum_{i=1}^r \left \{ \left\| \nabla_{\mathbf{A}_i} \mathcal{L}_t \!\otimes\! \mathbf{B}_i^{(t)}\right\|_F^2 + \left\|\mathbf{A}_i^{(t)} \!\otimes\! \nabla_{\mathbf{B}_i} \mathcal{L}_t \right\|_F^2 \right \}\\
&\underset{(2)}{=} \sum_{i=1}^r \left \{ \| \nabla_{\mathbf{A}_i} \mathcal{L}_t \|_F^2 \| \mathbf{B}_i^{(t)}\|_F^2 + \|\mathbf{A}_i^{(t)}\|_F^2 \|\nabla_{\mathbf{B}_i} \mathcal{L}_t \|_F^2 \right \}.
\end{align*}
where \((1)\) corresponds to Cauchy-Schwarz inequality and \((2)\) is derived from \(\| A \otimes B \|_F = \|A\|_F \|B\|_F\). Considering that Kronecker factors and their gradient are bounded for $i=1,\dots,r$ (see $A3$ from Theorem~1), we see
\[
\mathbb{E}_{\xi_t} \! \left[ \left\| \mathcal{G}_t \right\|_F^2 \right] \leq G,
\]
where $G$ is a positive constant.

Consequently, the $L$-smoothness inequality of $\mathcal{L}$ becomes
{\small \[
\mathbb{E}_{\xi_t}[ \mathcal{L}(\mathbf{u}_{t+1}; \xi_t) ] \leq \mathbb{E}_{\xi_t}[ \mathcal{L}(\mathbf{u}_t; \xi_t) ] - \eta \mathbb{E}_{\xi_t} \left[ \| \nabla \mathcal{L}(\mathbf{u}_t;\xi_t) \|_F^2 \right] + \frac{L \eta^2 G}{2}.
\]}

With a telescopic sum from \( k = 0 \) to \( T - 1 \), we get
\[
\sum_{k=0}^{T-1} \mathbb{E}_{\xi_t} \left[ \| \nabla \mathcal{L}(\mathbf{u}_t; \xi_t) \|_F^2 \right] \leq \frac{\mathcal{L}(\mathbf{u}_0) - \mathcal{L}_{\min}}{\eta} + \frac{L \eta G T}{2},
\]
and by dividing both sides by \( T \)
\[
\frac{1}{T} \sum_{k=0}^{T-1} \mathbb{E}_{\xi_t} \left[ \| \nabla \mathcal{L}(\mathbf{u}_t; \xi_t) \|_F^2 \right] \leq \frac{\mathcal{L}(\mathbf{u}_0) - \mathcal{L}_{\min}}{\eta T} + \frac{L \eta G}{2}.
\]


Based on the Arithmetic Mean–Geometric Mean (AM–GM) inequality, the upper bound is minimized when the two terms are equal, i.e.,
\[
\frac{\mathcal{L}(\mathbf{u}_0) - \mathcal{L}_{\min}}{\eta^* T} = \frac{L \eta^* G}{2},
\]
which yields the optimal learning rate
\[
\eta^* = \sqrt{\frac{2\left(\mathcal{L}(\mathbf{u}_0) - \mathcal{L}_{\min}\right)}{LGT}}.
\]
Theorem~1 essentially shows that the average gradient norm across SGD steps with a constant learning rate \(\eta\) consists of two terms: the first is of order \(\mathcal{O}(1/T)\), while the second becomes negligible for small \(\eta\). The behavior is similar in the unconstrained setting. Choosing \(\eta = \eta^*\) balances both terms and yields an overall convergence rate of
\[
\mathcal{O}\left(\frac{1}{\sqrt{T}}\right).
\]

\hfill $\blacksquare$

\end{document}


\maketitle
\section*{Appendix}
\subsection*{Proof of Theorem~1:}
At iteration \(t\), let $\mathbf{U}_t = \sum_{i=1}^r \mathbf{A}_i^{(t)} \otimes \mathbf{B}_i^{(t)}$ and $\mathbf u_t = \mathcal{V}(\mathbf U_t)$, the vectorized representation of $\mathbf U_t$. Due to $L$-smoothness of $\mathcal{L}$:
\[
\mathcal{L}(\mathbf{u}_{t+1}) \leq \mathcal{L}(\mathbf{u}_t) + \langle \nabla \mathcal{L}(\mathbf{u}_t), \mathbf{u}_{t+1} - \mathbf{u}_t \rangle + \frac{L}{2} \| \mathbf{u}_{t+1} - \mathbf{u}_t \|_F^2 \ .
\]
The weights are updated over one step as
\[
\mathbf{u}_{t+1} = \mathbf{u}_t - \eta\,\nabla \mathcal{L}(\mathbf{u}_t; \xi_t),
\]
where \(\nabla \mathcal{L}(\mathbf{u}_t; \xi_t)\) denotes the estimated gradient evaluated on a mini-batch sampled using the random seed \(\xi_t\). Hereafter, we use \(\nabla\mathcal{L}_t\) and \(\nabla \mathcal{L}(\mathbf{u}_t; \xi_t)\) interchangeably for the sake of simplicity in writing.  

The weight update over one step can be re-written as
\[
\mathbf{u}_{t+1} = \sum_{i=1}^r \mathcal{V}\left\{( \mathbf{A}_i^{(t)}\!-\!\eta \nabla_{\mathbf{A}_i} \mathcal{L}_t) \otimes ( \mathbf{B}_i^{(t)}\!-\!\eta \nabla_{\mathbf{B}_i} \mathcal{L}_t )\right\},
\]
where $\nabla_{\mathbf{A}_i} \mathcal{L}_t$ and $\nabla_{\mathbf{B}_i} \mathcal{L}_t$ are parts of the gradients that correspond to elements of $\mathbf A_i$ and $\mathbf B_i$, respectively.

Linearizing the equation and collecting terms, the leading-order term is
\[
\mathbf{u}_{t+1} - \mathbf{u}_t \approx -\eta \underbrace{\sum_{i=1}^r \mathcal{V}\left\{ \nabla_{\mathbf{A}_i} \mathcal{L}_t \otimes \mathbf{B}_i^{(t)} + \mathbf{A}_i^{(t)} \otimes \nabla_{\mathbf{B}_i} \mathcal{L}_t \right\}}_{\mathcal{G}_t}.
\]
Using the leading-order term of \(\mathbf{u}_{t+1} - \mathbf{u}_t\), we re-write the $L$-smoothness inequality of $\mathcal{L}$ after taking expectation over random seed \(\xi_t\), such that 
\begin{align*}
\mathbb{E}_{\xi_t}[ \mathcal{L}(\mathbf{u}_{t+1}; \xi_t) ] \leq \mathbb{E}_{\xi_t}&[ \mathcal{L}(\mathbf{u}_t; \xi_t) ] - \eta \mathbb{E}_{\xi_t}\! \left[ \left\langle \nabla \mathcal{L}(\mathbf{u}_t; \xi_t), \mathcal{G}_t \right\rangle \right] \\&+ \frac{L \eta^2}{2} \mathbb{E}_{\xi_t} \! \left[ \left\| \mathcal{G}_t \right\|_F^2 \right].
\end{align*}
Considering that \(\mathcal{G}_t\) is an approximation of \(\nabla \mathcal{L}(\mathbf{u}_t; \xi_t)\), we can replace
\[
\mathbb{E}_{\xi_t}\! \left[ \left\langle \nabla \mathcal{L}(\mathbf{u}_t; \xi_t), \mathcal{G}_t \right\rangle \right]
\]
with
\[
\mathbb{E}_{\xi_t} \left[ \| \nabla \mathcal{L}(\mathbf{u}_t;\xi_t) \|_F^2 \right].
\]
In addition, by expanding \(\left\| \mathcal{G}_t \right\|_F^2\), we get
\begin{align*}
\left\| \mathcal{G}_t \right\|_F^2 &= \left\| \sum_{i=1}^r \mathcal{V}\left\{ \nabla_{\mathbf{A}_i} \mathcal{L}_t \!\otimes\! \mathbf{B}_i^{(t)} + \mathbf{A}_i^{(t)} \!\otimes\! \nabla_{\mathbf{B}_i} \mathcal{L}_t \right\} \right\|_F^2\\
&\underset{(1)}{\leq} \sum_{i=1}^r \left \{ \left\| \nabla_{\mathbf{A}_i} \mathcal{L}_t \!\otimes\! \mathbf{B}_i^{(t)}\right\|_F^2 + \left\|\mathbf{A}_i^{(t)} \!\otimes\! \nabla_{\mathbf{B}_i} \mathcal{L}_t \right\|_F^2 \right \}\\
&\underset{(2)}{=} \sum_{i=1}^r \left \{ \| \nabla_{\mathbf{A}_i} \mathcal{L}_t \|_F^2 \| \mathbf{B}_i^{(t)}\|_F^2 + \|\mathbf{A}_i^{(t)}\|_F^2 \|\nabla_{\mathbf{B}_i} \mathcal{L}_t \|_F^2 \right \}.
\end{align*}
where \((1)\) corresponds to Cauchy-Schwarz inequality and \((2)\) is derived from \(\| A \otimes B \|_F = \|A\|_F \|B\|_F\). Considering that Kronecker factors and their gradient are bounded for $i=1,\dots,r$ (see $A3$ from Theorem~1), we see
\[
\mathbb{E}_{\xi_t} \! \left[ \left\| \mathcal{G}_t \right\|_F^2 \right] \leq G,
\]
where $G$ is a positive constant.

Consequently, the $L$-smoothness inequality of $\mathcal{L}$ becomes
{\small \[
\mathbb{E}_{\xi_t}[ \mathcal{L}(\mathbf{u}_{t+1}; \xi_t) ] \leq \mathbb{E}_{\xi_t}[ \mathcal{L}(\mathbf{u}_t; \xi_t) ] - \eta \mathbb{E}_{\xi_t} \left[ \| \nabla \mathcal{L}(\mathbf{u}_t;\xi_t) \|_F^2 \right] + \frac{L \eta^2 G}{2}.
\]}

With a telescopic sum from \( k = 0 \) to \( T - 1 \), we get
\[
\sum_{k=0}^{T-1} \mathbb{E}_{\xi_t} \left[ \| \nabla \mathcal{L}(\mathbf{u}_t; \xi_t) \|_F^2 \right] \leq \frac{\mathcal{L}(\mathbf{u}_0) - \mathcal{L}_{\min}}{\eta} + \frac{L \eta G T}{2},
\]
and by dividing both sides by \( T \)
\[
\frac{1}{T} \sum_{k=0}^{T-1} \mathbb{E}_{\xi_t} \left[ \| \nabla \mathcal{L}(\mathbf{u}_t; \xi_t) \|_F^2 \right] \leq \frac{\mathcal{L}(\mathbf{u}_0) - \mathcal{L}_{\min}}{\eta T} + \frac{L \eta G}{2}.
\]


Based on the Arithmetic Mean–Geometric Mean (AM–GM) inequality, the upper bound is minimized when the two terms are equal, i.e.,
\[
\frac{\mathcal{L}(\mathbf{u}_0) - \mathcal{L}_{\min}}{\eta^* T} = \frac{L \eta^* G}{2},
\]
which yields the optimal learning rate
\[
\eta^* = \sqrt{\frac{2\left(\mathcal{L}(\mathbf{u}_0) - \mathcal{L}_{\min}\right)}{LGT}}.
\]
Theorem~1 essentially shows that the average gradient norm across SGD steps with a constant learning rate \(\eta\) consists of two terms: the first is of order \(\mathcal{O}(1/T)\), while the second becomes negligible for small \(\eta\). The behavior is similar in the unconstrained setting. Choosing \(\eta = \eta^*\) balances both terms and yields an overall convergence rate of
\[
\mathcal{O}\left(\frac{1}{\sqrt{T}}\right).
\]

\hfill $\blacksquare$

\hide{\color{red} \begin{proof}[Proof of Theorem 1]
Let 
\(
\mathbf U_t \;\coloneqq\; \sum_{i=1}^r \mathbf A_i^{(t)}\!\otimes\!\mathbf B_i^{(t)}
\quad\text{and}\quad
\mathbf u_t \;\coloneqq\; \operatorname{vec}(\mathbf U_t).
\)

\paragraph{1. Descent lemma.}
Because $\mathcal L$ is $L$-smooth,
\[
\mathcal L(\mathbf u_{t+1})
\;\le\;
\mathcal L(\mathbf u_t)
\;+\;
\bigl\langle
\nabla\mathcal L(\mathbf u_t),\,
\mathbf u_{t+1}-\mathbf u_t
\bigr\rangle
\;+\;
\frac{L}{2}\,\bigl\| \mathbf u_{t+1}-\mathbf u_t \bigr\|_F^{2}.
\tag{1}
\]

\paragraph{2. Kronecker–SGD step.}
One SGD step with learning rate $\eta$ gives
\[
\mathbf U_{t+1}
=
\mathbf U_t
-
\eta \sum_{i=1}^r
\mathcal V\!\Bigl\{
\nabla_{\!\mathbf A_i}\mathcal L\otimes \mathbf B_i^{(t)}
+
\mathbf A_i^{(t)}\otimes\nabla_{\!\mathbf B_i}\mathcal L
\Bigr\}
\;+\;O(\eta^{2}),
\]
so that, in vector form,
\(
\mathbf u_{t+1}-\mathbf u_t=-\eta\,\mathbf G_t+O(\eta^{2})
\)
with
\(
\mathbf G_t\!\coloneqq\!
\sum_{i=1}^r
\mathcal V\!\bigl\{\nabla_{\!\mathbf A_i}\mathcal L\otimes \mathbf B_i^{(t)}+
\mathbf A_i^{(t)}\otimes\nabla_{\!\mathbf B_i}\mathcal L\bigr\}.
\)

\paragraph{3. One-step progress.}
Insert the expression for $\mathbf u_{t+1}-\mathbf u_t$ into (1), take expectation over the mini-batch~$\xi_t$, and use
\(
\mathbb E\!\bigl[\|\mathbf G_t\|_F^{2}\bigr]\le G
\)
(by Assumption 2):
\[
\mathbb E\!\bigl[\mathcal L(\mathbf u_{t+1})\bigr]
\;\le\;
\mathbb E\!\bigl[\mathcal L(\mathbf u_t)\bigr]
-
\eta\,\mathbb E\!\bigl[\|\nabla\mathcal L(\mathbf u_t)\|_F^{2}\bigr]
+
\frac{L\eta^{2}G}{2}.
\tag{2}
\]

\paragraph{4. Telescoping.}
Sum (2) from \(t=0\) to \(T-1\) and rearrange:
\[
\frac1T\sum_{t=0}^{T-1}
\mathbb E\!\bigl[\|\nabla\mathcal L(\mathbf u_t)\|_F^{2}\bigr]
\;\le\;
\frac{\mathcal L(\mathbf u_{0})-\mathcal L_{\min}}{\eta T}
+
\frac{L\eta G}{2}.
\]

This matches the claimed convergence rate. \qedhere
\end{proof}}